\documentclass[letterpaper]{article} 
\usepackage{aaai25}  
\usepackage{times}  
\usepackage{helvet}  
\usepackage{courier}  
\usepackage[hyphens]{url}  
\usepackage{graphicx} 
\usepackage{natbib}  
\usepackage{caption} 
\usepackage{algorithm}
\usepackage{algorithmic}
\usepackage{amsmath}
\usepackage{amssymb}
\usepackage{color}
\usepackage{multirow} 
\usepackage{newfloat}
\usepackage{listings}
\DeclareCaptionStyle{ruled}{labelfont=normalfont,labelsep=colon,strut=off} 
\floatstyle{ruled}
\newfloat{listing}{tb}{lst}{}
\floatname{listing}{Listing}
\title{Hierarchically-Structured Open-Vocabulary Indoor Scene Synthesis  
\par with Pre-trained Large Language Model}
\author{
    Weilin Sun\textsuperscript{\rm 1},
    Xinran Li\textsuperscript{\rm 1},
    Manyi Li\textsuperscript{\rm 1}\thanks{Corresponding Author.},
    Kai Xu\textsuperscript{\rm 2}\footnotemark[1],
    Xiangxu Meng\textsuperscript{\rm 1},
    Lei Meng\textsuperscript{\rm 1, 3}\footnotemark[1]
}
\affiliations{
    \textsuperscript{\rm 1}School of Software, Shandong University, China\\
    \textsuperscript{\rm 2} School of Computer Science, National University of Defense Technology, China\\
    \textsuperscript{\rm 3} Shandong Research Institute of Industrial Technology, China\\
    \{sunweilin, xinranli\}@mail.sdu.edu.cn, manyili@sdu.edu.cn, kevin.kai.xu@gmail.com, \{mxx, lmeng\}@sdu.edu.cn, 
}

\usepackage{bibentry}

\begin{document}

\maketitle

\begin{abstract}
Indoor scene synthesis aims to automatically produce plausible, realistic and diverse 3D indoor scenes, especially given arbitrary user requirements. Recently, the promising generalization ability of pre-trained large language models (LLM) assist in open-vocabulary indoor scene synthesis. However, the challenge lies in converting the LLM-generated outputs into reasonable and physically feasible scene layouts. In this paper, we propose to generate hierarchically structured scene descriptions with LLM and then compute the scene layouts. Specifically, we train a hierarchy-aware network to infer the fine-grained relative positions between objects and design a divide-and-conquer optimization to solve for scene layouts. The advantages of using hierarchically structured scene representation are two-fold. First, the hierarchical structure provides a rough grounding for object arrangement, which alleviates contradictory placements with dense relations and enhances the generalization ability of the network to infer fine-grained placements. Second, it naturally supports the divide-and-conquer optimization, by first arranging the sub-scenes and then the entire scene, to more effectively solve for a feasible layout. We conduct extensive comparison experiments and ablation studies with both qualitative and quantitative evaluations to validate the effectiveness of our key designs with the hierarchically structured scene representation. Our approach can generate more reasonable scene layouts while better aligned with the user requirements and LLM descriptions. We also present open-vocabulary scene synthesis and interactive scene design results to show the strength of our approach in the applications.
\end{abstract}

%

\section{Introduction}

\begin{figure}[t]
\centering
\includegraphics[width=\linewidth]{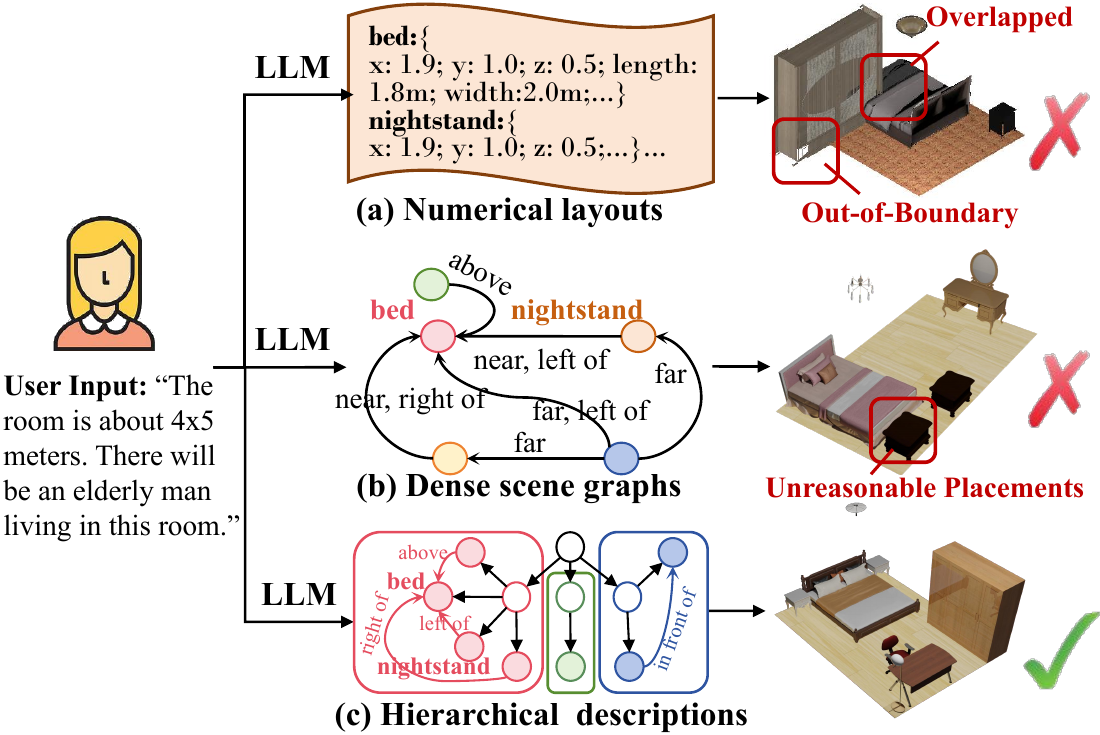}
\caption{Compared to LLM-generated (a) numerical layouts and (b) scene graphs with dense relations, we use (c) LLM-generated hierarchical scene descriptions, whose internal nodes represent functional areas with compact and generalizable prior, to generate more reasonable and physically feasible scene layouts aligned with the descriptions.}
\label{pipeline}
\end{figure}

Indoor scene design requires a comprehensive consideration of space partition, functional arrangement, and aesthetic creativity to determine the object selection and placement to form the scene layout. There has been a vast amount of research on indoor scene synthesis, ranging from layout optimization~\cite{Yu2011MakeIH, Merrell2011InteractiveFL, Qi2018HumanCentricIS} to various conditional scene synthesis~\cite{Wang2018DeepCP, Paschalidou2021ATISSAT, Gao2023SceneHGNHG, Tang2023DiffuSceneSG}. The goal is to automatically produce plausible, realistic, and diverse 3D indoor scenes, especially
given arbitrary user requirements. However, due to the complexity of indoor scenes, most of them are limited within the scope of training data and cannot be generalized to arbitrary conditions.

Some pioneer works~\cite{Feng2023LayoutGPTCV, Wen2023AnyHomeOG, Yang2023HolodeckLG} make use of the promising generalization ability of pre-trained large language models (LLM) to address the open-vocabulary scene synthesis task, where the LLM is responsible for interpreting any textual requirement into detailed scene configurations. The challenge lies in obtaining reasonable and physically feasible scene layouts from LLM outputs. Directly using LLM to output numerical layouts~\cite{Feng2023LayoutGPTCV} causes unreliable results with heavy object overlap and out-of-boundary, since LLM fails to understand the spatial relationship with numerical layouts. On the other hand, LLM shows good performance in generating detailed text descriptions of various scenes, but still require an approach to convert the textual descriptions into numerical layouts while maintaining the generalization of the entire pipeline. The existing methods~\cite{Wen2023AnyHomeOG, Yang2023HolodeckLG} have to pre-define textual phrases and numerical rules for several types of spatial relations to obtain the layouts. However, dense spatial relations often lead to incompatible object arrangements while coarse relations fail to capture diverse spatial placements, causing misalignment between LLM-generated configuration and the results. 

\begin{figure}
\centering
\includegraphics[width=\linewidth]{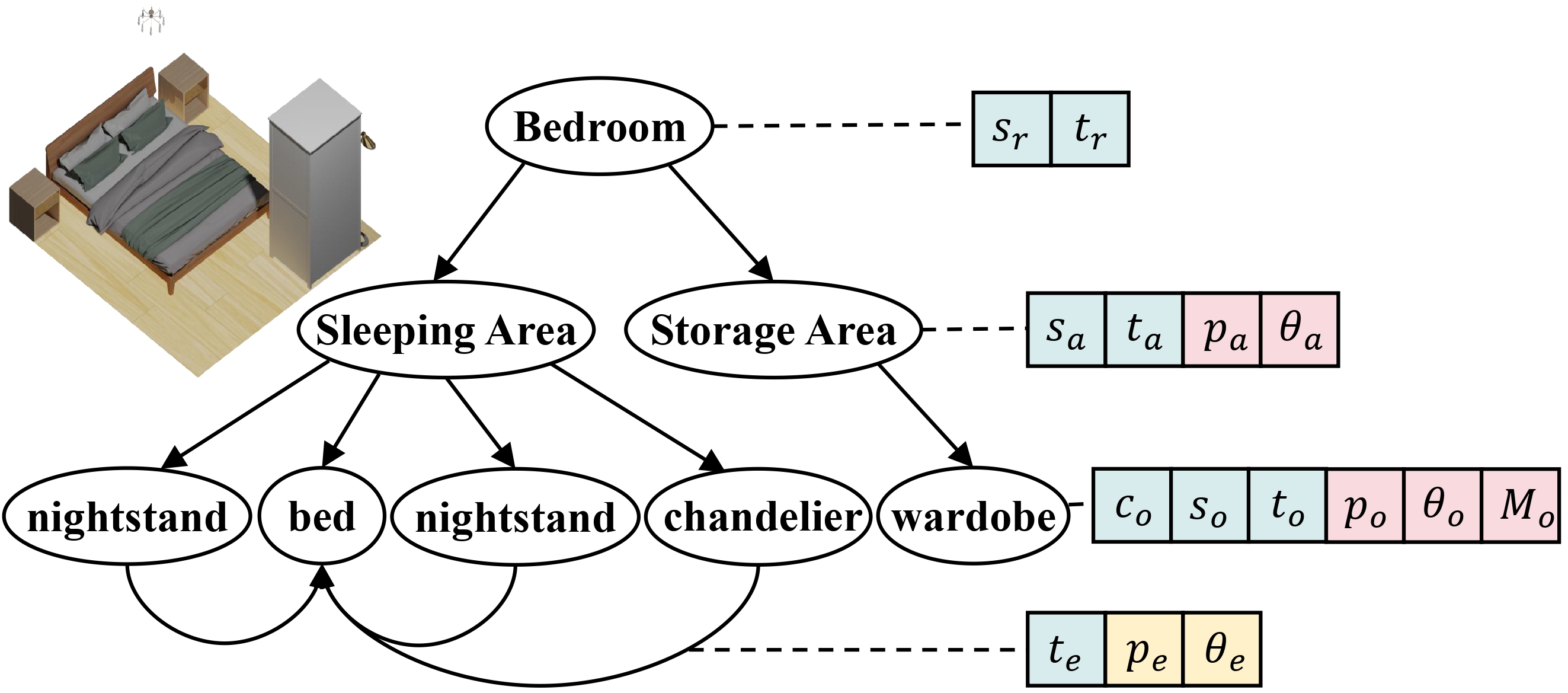}
\caption{Our three-level hierarchical scene structure with the functional area as internal nodes.}
\label{fig:hierarchy}
\end{figure}

In this paper, we propose to use hierarchical scene descriptions as the intermediate representation in the LLM-assisted scene synthesis pipeline. The hierarchical structure has three levels, with the entire scene as root node, functional areas as internal nodes, and objects as leaf nodes, as illustrated in Figure~\ref{fig:hierarchy}. Our approach contains three stages. First, we prompt the pre-trained LLM to generate the hierarchical structure with text descriptions. Second, we train a hierarchy-aware network to further infer the fine-grained relative placements between objects with textual spatial relations. Taking the hierarchical structure as grounding, the network can infer reasonable relative placements in an open-vocabulary setting. Third, we develop a divide-and-conquer optimization, which optimizes each functional area separately and then arranges them to form the entire scene, to solve for the physically feasible scene layouts effectively.

The advantages of using hierarchically structured scene representation are two-fold. First, the hierarchical structure provides a rough grounding for object arrangement, which alleviates contradictory placements with dense relations and enhances the generalization ability of the network to infer fine-grained placements. Second, it naturally supports the divide-and-conquer optimization to more effectively solve for a feasible layout that matches with the LLM-generated descriptions. We perform extensive comparison experiments and ablation studies with both qualitative and quantitative evaluations. Our approach generates more reasonable and physically feasible scenes that align better with the user requirements and LLM arrangements. In addition, we show the results of open-vocabulary scene synthesis and interactive scene design as practical applications of our approach.

Our contributions are summarized as follows:
\begin{itemize}
\item We propose an LLM-assisted hierarchically-structured scene synthesis pipeline, which uses a three-level hierarchical structure to infer reasonable object arrangements.
\item We develop the hierarchy-aware network and the divide-and-conquer optimization, which takes advantage of the hierarchical structure for effective layout synthesis.
\item We conducted extensive comparison and ablation study experiments to validate the effectiveness of our approach, as well as two applications to show its practical usage.
\end{itemize}

\section{Related Work}

\noindent \textbf{Indoor Scene Synthesis.} The common practice is to produce a set of objects and their placements~\cite{Patil2023AdvancesID}, i.e. scene layouts. Early works rely on the pre-defined rules~\cite{Yu2011MakeIH, Merrell2011InteractiveFL, Ma2016Actiondriven3I, Fu2017AdaptiveSO, Fu2020HumancentricMF} to generate interpretable and feasible scene layouts. To further capture diverse spatial arrangement, the data-driven approaches~\cite{fisher2012example, Qi2018HumanCentricIS, xu2014organizing, ma2018language, Sun2022SequentialFO, Sun2024SequentialSA} learn the object relationship from datasets~\cite{fu20213d, song2017semantic}. The researchers have developed all kinds of networks to learn the scenes represented as different data structures, including sequences~\cite{Wang2018DeepCP}, graphs~\cite{Zhou2019SceneGraphNetNM, Wang2019PlanITPA}, hierarchies~\cite{li2019grains, Gao2023SceneHGNHG}, sets~\cite{Paschalidou2021ATISSAT, Wei2023LEGONetLR, Tang2023DiffuSceneSG, zhai2024echoscene}, etc. However, due to the inherent complexity, it is difficult to capture the essential relationship from the observed layouts and generalize to other categories.

Incorporating additional knowledge enhances scene prior learning.  Graph-to-3D~\cite{Dhamo2021Graphto3DEG} and CommonScenes~\cite{zhai2024commonscenes} synthesize the layouts and object shapes for coherent scenes. Some methods~\cite{ye2022scene, yi2023mime} take human motion trajectory as conditions to populate the objects. Haisor~\cite{sun2024haisor} uses reinforcement learning with human interaction and space area consideration for scene synthesis. External expert knowledge~\cite{leimer2022layoutenhancer, yang2024learning} can also be incorporated during training to enhance network performance. These works refer to the same observation that indoor scene synthesis involves a comprehensive consideration of space partition, functional arrangement, and aesthetic creativity, thus requiring a generative model with extensive knowledge.

\noindent \textbf{Text-to-Scene Synthesis.} The challenges include semantic understanding of user requirements and scene synthesis. Given a natural language description, early works~\cite{ChangSM14, ChangMSPM15, ChangESM17, SavvaCA17, MaPFLPHYTGZ18, Yang2021} parse the input as scene templates, where the nodes represent objects and edges for spatial relations, and then sample the corresponding object models and placements. With the development of deep learning, some works~\cite{Paschalidou2021ATISSAT, Tang2023DiffuSceneSG, Dinh2024} take latent vectors such as textual embeddings as conditions, and train conditional scene synthesis model to output scenes. However, these works rely on detailed descriptions as input to specify the setting of target scenes, e.g. "there is a desk and there is a notepad on the desk", rather than reasoning the scene configuration from abstract instructions. Moreover, generalizing to diverse scene categories and open-vocabulary settings is a long-standing problem. 

\begin{figure*}
\centering
\includegraphics[width=\linewidth]{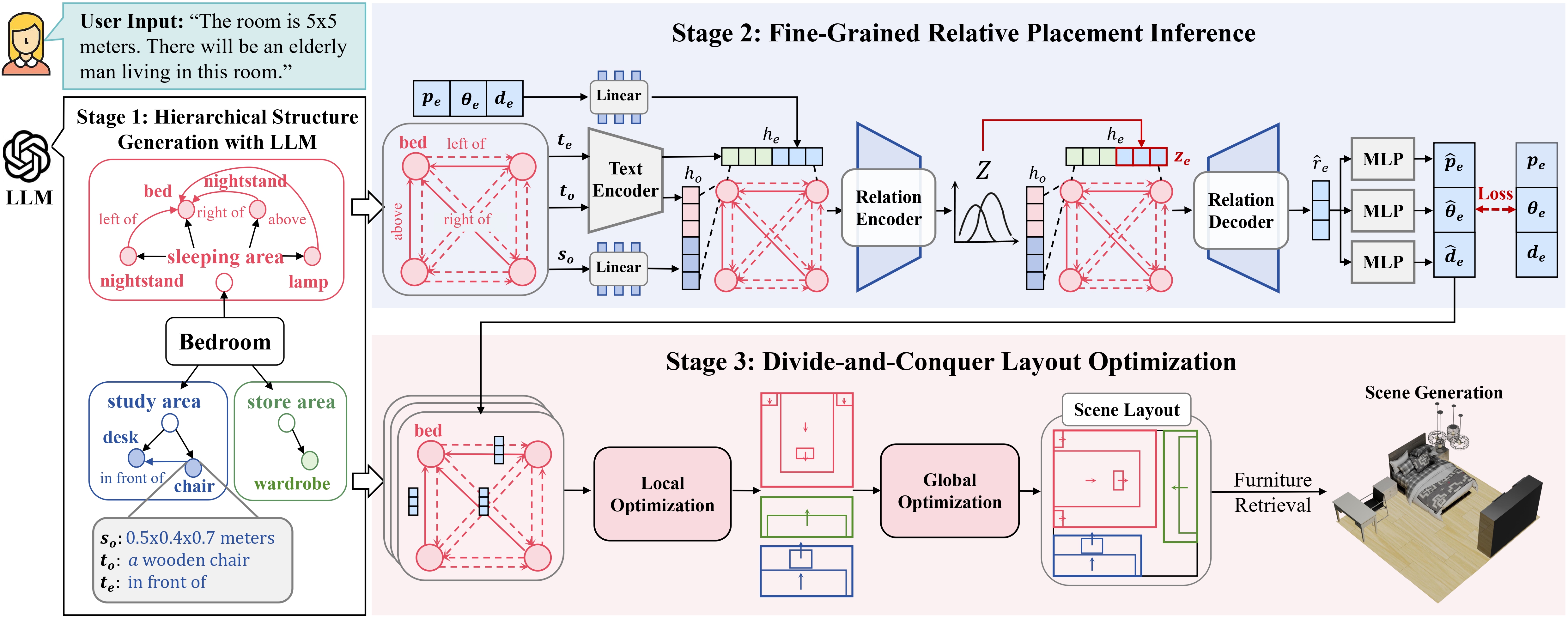}
\caption{Our hierarchically-structured scene synthesis pipeline involves three stages. First, given a user requirement, we prompt the LLM to generate a hierarchical structure with text descriptions. Second, we train a hierarchy-aware network to infer the fine-grained relative placements between objects. Third, we use a divide-and-conquer optimization algorithm to arrange each functional area separately and then organize them into the entire scene. $z_e$ indicates the random sampling from the Gaussian distribution $Z$, and $\hat{r}_e$ indicates the relative position information after decoding.}
\label{pipeline}
\end{figure*}


\noindent \textbf{LLM-Assisted Scene Synthesis.} Recent works have investigated utilizing the pre-trained LLMs to handle the multi-objects in the scenes, most of which focus on the 2D layout for controllable scene image synthesis~\cite{lian2023llm, gani2023llm, nie2024compositional}. LayoutGPT~\cite{Feng2023LayoutGPTCV} retrieves scene layouts in CSS format, with LLMs outputting numerical bounding boxes for each object. However, due to the lack of spatial reasoning ability, LLM cannot handle the complex relationships of 3D scenes, causing heavy object overlap and out-of-boundary problems.

Some others use LLM to generate a textual scene description and convert it to 3D scenes. Aladdin~\cite{Huang2023AladdinZH} introduced a pipeline to sample and generate 3D textured assets from an abstract description and manually organize them to construct a scene. Recent works~\cite{Wen2023AnyHomeOG, Yang2023HolodeckLG} require LLM to describe the object relations using pre-defined atom relations, which are interpreted as fixed relative positions between objects and refined using a rule-based optimization algorithm. However, our experiments show that defining a compact yet informative atom relations is difficult. Dense and detailed object relations are able to provide precise spatial arrangement but often cause self-contradiction in the LLM outputs, while a sparse set of coarse relations leads to coherent arrangements but fails to capture the diverse spatial placements among objects. 

\section{Problem Formulation}

\noindent \textbf{Problem.} Given an arbitrary textual description $t$ as the input condition, e.g. "a small bedroom for a young student", our goal is to generate plausible, realistic, and physically feasible scenes corresponding to the requirement. The generated scene includes the selected 3D models and the arranged placements of the objects therein, i.e. $S=\{(M_i, P_i)|i=1,...,n\}$, where $M_i$ and $P_i$ denote the selected 3D model and spatial placement respectively for the $i$th object. The placement of each object contains its center position coordinates, orientation angle, and size.

\noindent \textbf{Hierarchical Scene Representation.} 
We propose to use the hierarchical scene representation throughout our pipeline. It is a three-level hierarchical structure, as shown in Figure~\ref{fig:hierarchy}. The first level is the root node representing the entire scene, the second level is the internal nodes each representing a rectangular functional area, and the third level is the leaf nodes representing the objects belonging to the corresponding area. The nodes are connected with two types of edges,
i.e. the parent-child relation indicating the hierarchical structure and the pairwise relation between objects to represent
their spatial relationship. Specifically, to reduce the redundancy, we set one anchor object for each functional area and only allow the pairwise relations between the anchor object and other objects belonging to the same functional area.

Each node in the hierarchy contains some attributes. Assuming axis-aligned rectangular floorplans for the scenes, the root node $r$ has a size attribute $s_r$ and a text description $t_r$ of the scene, i.e. $r = \{t_r, s_r\}$. Each internal node $a$, which represents an axis-aligned functional area, carries the text description $t_a$, the size attributes $s_a$, as well as a center position $p_a$ and an orientation $\theta_a$, i.e. $a = \{t_a, s_a, p_a, \theta_a\}$. The position is a 2D coordinate while the orientation is a binary value representing either horizontal or vertical direction. Each object node $o$ contain the text description $t_o$, the category label $c_o$, the corresponding 3D model $M_o$, as well as the size $s_o$, center position $p_o$, orientation $\theta_o$ of the oriented bounding box of the object, i.e. $o = \{t_o, c_o, M_o, s_o, p_o, \theta_o\}$. Note that the size attributes are 2D vectors for room and functional areas, but 3D vectors for objects, since we also care about their heights. In addition, the pairwise spatial relationship $e$ stores the coarse text description $t_e$ such as "in front of" and the fine-grained relative placement coordinates including position $p_e$ and orientation $\theta_e$ of one object w.r.t. the anchor object, i.e. $e = \{t_e, p_e, \theta_e\}$.
\section{Hierarchical Scene Synthesis with LLM}

Given a text description $t_r$ and the scene size $s_r$ as conditions, our approach is composed of three stages, as illustrated in Figure~\ref{pipeline}. First, we prompt the pre-trained LLM to generate the hierarchical structure with text descriptions. Second, we train a hierarchy-aware graph neural network to infer the relative placement coordinates between objects.
Third, we design a divide-and-conquer optimization which optimizes the sub-layout for each functional area and then arranges their placements to form the entire scene.  

\subsection{Hierarchical Structure Generation with LLM} 
Given the user requirement, the pre-trained LLM takes the constructed prompt as input and outputs structured text to describe the hierarchical scene representation, including the node attributes. The key challenge is to generate reasonable and informative spatial relations to specify the scene layout.

Although existing works define dense object relations to describe layouts, the more detailed the descriptions are, the more incorrect or self-contradictory results they make, due to the lack of spatial reasoning ability of the LLM. Therefore, in our approach, we require the LLM to generate a hierarchical structure to ground the objects and only the spatial relations between objects belonging to the same area, only to roughly specify their arrangements.

We construct the input prompt with three components: 1) a description of the LLM's role and task, including a brief definition of the hierarchical structure with the meaning of the nodes and connections; 2) a description of the preferred data format and pre-defined constraints, including the types of functional areas, possible anchor objects, and spatial relations; and 3) an example of a simple scene in the preferred format and the specific user requirements. We don't require the example to be selected corresponding to the user requirement, but only to demonstrate the output format. In this stage, the LLM generates textual descriptions and size attributes of the functional areas and objects, as well as the textual descriptions of spatial relations.

\subsection{Fine-Grained Relative Placement Inference}
We propose a hierarchy-aware graph neural network to infer the fine-grained relative placements between correlated objects. The relative placements within each functional area exhibit a more compact and generalizable prior, allowing us to train a network to infer the placements for various scenes.

As illustrated in Figure~\ref{pipeline}, given the LLM-generated hierarchy, we construct the input graph $G=(O,E)$ with the nodes as objects and edges connecting all objects belonging to the same functional area. Although the input includes all objects in the scene, the functional areas are isolated from each other. We use Linear embeddings for the object sizes $s_o$ and the ground truth relative placement coordinates $[p_e, \theta_e, d_e]$, where $d_e$ is a binary indicator of the alignment between two objects, and the pre-trained CLIP text encoder~\cite{radford2021learningtransferablevisualmodels} for descriptions of objects $t_o$ and spatial relations $t_e$. They are organized as node features $h_o$ and edge features $h_e$, i.e.

\begin{equation}
\begin{aligned}
& h_o = [\mathrm{CLIP_t}(t_o), \mathrm{LINEAR(s_o)}], \\
& h_e = [\mathrm{CLIP_t}(t_e), \mathrm{LINEAR}(p_e, \theta_e, d_e)],
\end{aligned}
\end{equation}
which forms the contextual graph for the following network processing. Note that since we only have textual spatial relations between the anchor object and the others, we use all-zero vectors as the text embeddings for the edges without corresponding textual spatial relations (dotted arrows). 

We adopt the variational graph neural network~\cite{zhai2024commonscenes} for the contextual graph with the $h_o$ and $h_e$. Both the encoder and decoder are composed of several MLPs for 5 rounds of message passing, including $g_e^{(k)}$ for updating the edge features with connected node features in the $k$th round and $g_o^{(k)}$ for updating the node features with the 1-ring neighbor nodes, i.e. 
\begin{equation}
\begin{aligned}
h_{e_{i\xrightarrow{} j}}^{(k+1)}&=g_e^{(k)}(h_{o_i}^{(k)}, h_{e_{i\xrightarrow{} j}}^{(k)}, h_{o_j}^{(k)}) \\
h_{o_i}^{(k+1)} &=h_{o_i}^{(k)} + g_o^{(k)}(\mathrm{AVG}(h_{o_j}^{(k)}|o_j \in N_ \mathcal{G}(o_i))),
\end{aligned}
\end{equation}
where $e_{i\xrightarrow{} j}$ represents an edge connecting two objects $o_i$ and $o_j$, $N_\mathcal{G}(o_i)$ represents the set of neighbor nodes connected with object $o_i$. The encoder takes the contextual graph as input and outputs the graph with updated features, where the edge features (specifically the relative placement components of edge features, as shown in Figure~\ref{pipeline}) are parameterized as a Gaussian distribution. The decoder takes the updated graph as input and randomly samples from the Gaussian distribution. Finally, we use separate MLPs to decode the relative placement $[\hat{p}_e, \hat{\theta}_e, \hat{d}_e]$.

During training, we freeze the CLIP text encoder and update all other network layers. The loss function is 
\begin{equation}
L=L_{KL} + L_{ep} + L_{e\theta} + L_{ed},
\end{equation}
where $L_{KL}$ is the Kullback-Liebler divergence between the Gaussian distribution and
posterior distribution of the edge feature components. $L_{ep}$ is L1 loss on the relative positions $p_e$. $L_{e\theta}$ and $L_{ed}$ are cross-entropy loss on the discretized relative orientation angles and the binary alignment indicator.

\subsection{Divide-and-Conquer Layout Optimization} 

Given the hierarchical scene with the relative placements between correlated objects, we develop a divide-and-conquer optimization to solve for the final layout. Our solution includes a local optimization for each functional area and then a global optimization to organize the areas into scenes. This optimization produces reasonable and physically feasible layouts more effectively than a simple global optimization or iteratively optimizing each object's placements.

\begin{figure*}
\centering
\includegraphics[width=0.95\linewidth]{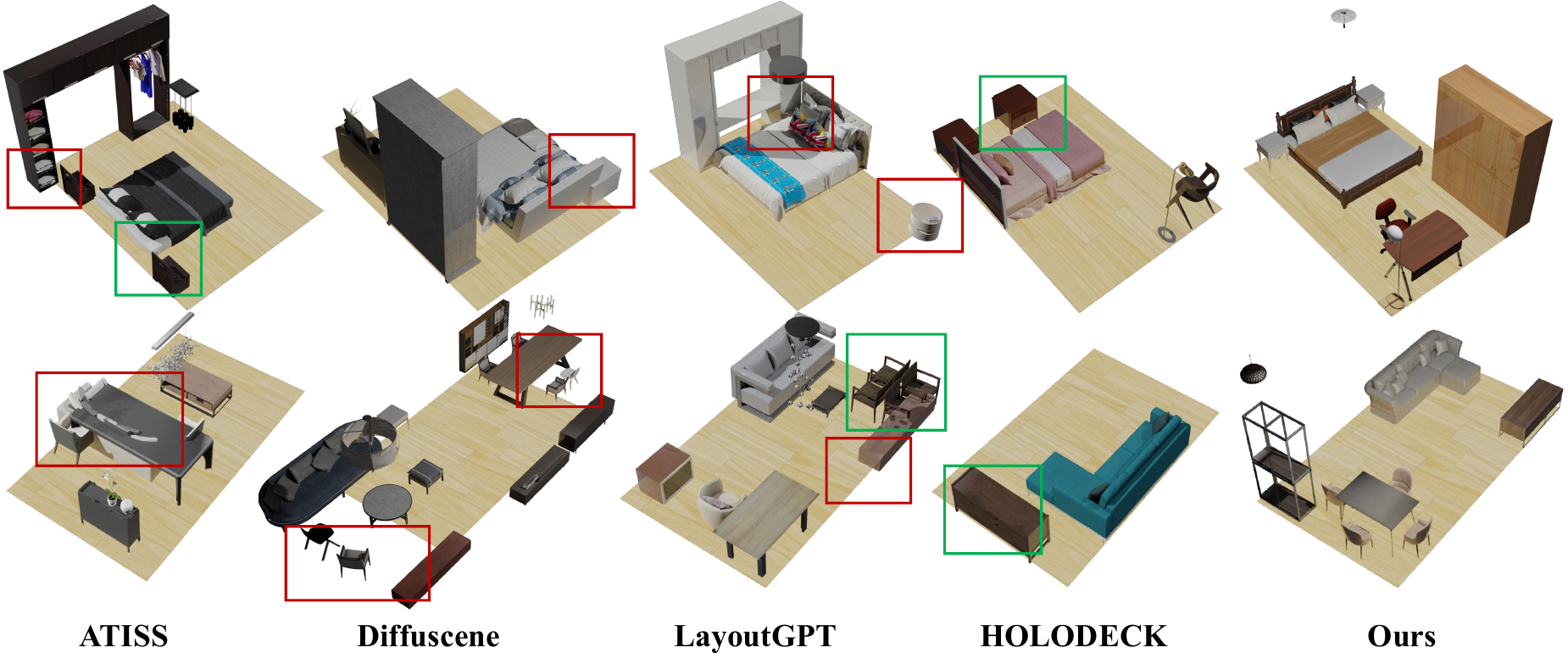}
\caption{The scenes generated from different approaches. The end-to-end data-driven approaches often produce infeasible object placements including overlap and out-of-boundary cases (red boxes). On the other hand, HOLODECK sometimes produces unreasonable results such as two nightstands on the same side of the bed (top row) or a tv stand on the left side of the sofa (second row). By contrast, our approach is able to more effectively produce reasonable and physically feasible scene layouts.}
\label{fig:comparison_topview}
\end{figure*}

\noindent \textbf{Local optimization.} For each functional area, we use local optimization to solve the object placements w.r.t. the bounding box of the functional area. The local optimization is formulated to minimize the objects' relative placements and those inferred by the network, with constraints to avoid object overlap and out-of-boundary, i.e.
\begin{equation}
\begin{aligned}
\min_{o'_i\in O} &  \sum_{o_i\in N_\mathcal{G}(o_a)} |\mathrm{REL}(o'_i, o'_a)-[p_{e_{i\xrightarrow{} a}},\theta_{e_{i\xrightarrow{} a}}]|, \quad \\
s.t. \quad & C_{overlap}(o'_i, o'_j), \quad \forall o_i,o_j \in A \\
& C_{OOB}(o'_i, s_a), \quad \forall o_i \in A 
\end{aligned}
\end{equation}
where $o'_i$ and $o'_a$ refers to the placements (center positions and orientations) w.r.t. the functional area of an object $o_i$ and the anchor object $o_a$, respectively. $\mathrm{REL}$ computes the relative placements between two objects and $[p_{e_{i\xrightarrow{} a}},\theta_{e_{i\xrightarrow{} a}}]$ is the relative positions between object $o_i$ and $o_a$ predicted by the network. $A$ represents the set of objects within the area. $C_{overlap}$ constrains the overlap between oriented bounding boxes of any two objects as small as possible, and $C_{OOB}$ aims to avoid the object boxes lying out of the area boundary, whose size $s_a$ is generated from pre-trained LLM in stage 1.

\noindent \textbf{Global optimization.} We then organize the areas to form scenes with global optimization. Each functional area takes the orientation of its anchor object as its own orientation. Based on observations in our daily life, the optimization is formulated to place the functional areas against the walls and far from each other with orientations pointing inside the scene, while avoiding object overlap and out-of-boundary:
\begin{equation}
\begin{aligned}
\min_{a_i} & \sum_{a_i} |\mathrm{D_w}(a_i, s_r)| - \sum_{a_i, a_j} |\mathrm{D_a}(a_i, a_j)|, \\
s.t. \quad & C_{overlap}(a_i, a_j), \quad \forall a_i,a_j \in S \\
& C_{OOB}(a_i, s_r), \quad \forall a_i \in S  
\end{aligned}
\end{equation}
where $a_i$ denotes area placement (center position and orientation). $D_w$ is the distance between back side of the area and the boundary of the scene. $D_a$ is the distance between two areas' bounding boxes. $S$ is the set of areas within the scene. $C_{overlap}$ and $C_{OOB}$ are same as in local optimization.

After the optimizations, we transform the coordinate systems to obtain object positions and orientations in the frame of scenes. Finally, we retrieve 3D object models from Objaverse~\cite{objaverseXL} and 3D-Front datasets~\cite{fu20213d} based on the CLIP scores, i.e. cosine similarity between object images and text embeddings. The object models are then scaled and placed according to the scene layouts.

\section{Experiment and Results}
\begin{table*}[]
\setlength{\tabcolsep}{12.5pt}
\begin{tabular}{lccccccccl}
\hline
\multirow{2}{*}{Models} & \multicolumn{4}{c}{Bedrooms}                                                                               & \multicolumn{4}{c}{Living Rooms}                                                                            \\ \cline{2-9} 
                        & {OOB↓} & {Overlap↓} & \multicolumn{2}{c}{KL Div.↓}  & {OOB↓} & {Overlap↓} & \multicolumn{2}{c}{KL Div.↓}   \\ \hline
ATISS~\cite{Paschalidou2021ATISSAT}                   & 0.48                  & 0.18                        & \multicolumn{2}{c}{0.19}          & 0.50                  & 0.34                      & \multicolumn{2}{c}{0.23}           \\ \hline
Diffuscene~\cite{Tang2023DiffuSceneSG}              & 0.77                  & 0.14                        & \multicolumn{2}{c}{0.29}          & 0.95                  & 0.30                      & \multicolumn{2}{c}{0.28}           \\ \hline
LayoutGPT~\cite{Feng2023LayoutGPTCV}               & 0.70                  & 0.16                        & \multicolumn{2}{c}{0.25}          & 0.64                  & 0.21                      & \multicolumn{2}{c}{0.36}           \\ \hline
HOLODECK~\cite{Yang2023HolodeckLG}                & 0.00                  & 0.00                        & \multicolumn{2}{c}{0.27}          & 0.00                  & 0.00                      & \multicolumn{2}{c}{0.34}           \\ \hline
Ours                    & \textbf{0.00}         & \textbf{0.00}              & \multicolumn{2}{c}{\textbf{0.09}}   & \textbf{0.00}                  & \textbf{0.00}                      & \multicolumn{2}{c}{\textbf{0.13}}           \\ \hline
\end{tabular}
\caption{Quantitative evaluation of the generated scene layouts of different approaches.}
\label{tab:comparison}
\end{table*}

\subsection{Experiment Settings}

\noindent\textbf{Dataset.} We conduct the comparison and ablation study on the 3D-Front dataset~\cite{fu20213d}. That is, we train the deep-learning-based approaches on this dataset and constrain the LLM-assisted methods to synthesize scenes with object categories within this dataset, for a fair comparison. Following LayoutGPT~\cite{Feng2023LayoutGPTCV}, we take the room category and floor, i.e. its width and height, as input conditions and filter out the scenes with irregular floors. The sizes of the training sets are 3397 and 690 for bedrooms and living rooms, while the corresponding test sets are 60 and 53.

\noindent\textbf{Metrics.} We evaluate the generated scenes from two perspectives. One is the physical feasibility of the scenes, estimated by the overlap between oriented bounding boxes and out-of-boundary metrics, i.e. overlap and OOB. The other is the reasonable organization of scenes, for which we select some common object pairs, i.e. bed-nightstand, table-chair, table-sofa, and measure the averaged KL-divergence between the relative placement distributions of the ground-truth scenes in the test sets and the generated scenes.

\noindent\textbf{Implementation.} We use GPT-4\cite{achiam2023gpt} for all the LLM-assisted approaches for the evaluation (the open-source LLaMA also works well with our approach). We train the hierarchy-aware neural network with 500 epochs using the Adam optimizer, where the batch size is 4 and the learning rate is 1e-4. The network is trained on the combination of the bedroom and living room training sets, which takes about 8 hours on a Nvidia 4090 GPU. Our divide-and-conquer optimization is implemented with the GUROBI solver~\cite{gurobi}. Our approach takes about 2 minutes to synthesize a scene with 8 objects.  



\subsection{Comparisons}

We compare with two types of state-of-the-art approaches, including those training deep neural networks from scratch, i.e. ATISS~\cite{Paschalidou2021ATISSAT} and DiffuScene~\cite{Tang2023DiffuSceneSG}, and LLM-assisted indoor scene synthesis, i.e. LayoutGPT~\cite{Feng2023LayoutGPTCV} and HOLODECK~\cite{Yang2023HolodeckLG}. We re-train the deep networks using their released code on the same train/test split. For the LLM-assisted approaches, we use their implementations of the pipelines and invoke the same version of GPT for the inference.

\noindent\textbf{Qualitative Evaluation.} Figure~\ref{fig:comparison_topview} presents generated scenes of different methods. The deep learning methods ATISS and DiffuScene generate results with reasonable placements of objects. But the networks are not guaranteed to ensure the physical feasibility of the scene layouts and sometimes cause object overlap and out of the floor boundary. LayoutGPT, which uses in-context learning to infer numerical layouts based on the demonstrated examples, generates many incorrect orientations and positions. HOLODECK produces relatively better results in terms of physical feasibility, but some objects are not placed in the optimal position as specified by the LLM. By contrast, our approach is able to produce more reasonable and feasible scene layouts.

\noindent\textbf{Quanlitative Evaluation.} Table~\ref{tab:comparison} validates the observations from the visual results. Among all the methods, we achieve the best in terms of both the physical feasibility (overlap and OOB) and the reasonable relative placements (KL Div.). It is interesting to see that the data-driven approaches are good at objects' relative positions and LLM-assisted optimization, i.e. HOLODECK, obtains more feasible results, while ours takes the merit of both and won the best on all the metrics.

\begin{table}
\centering
\setlength{\tabcolsep}{8.5pt}
\begin{tabular}{ccccc}
\hline
\multirow{2}{*}{Models}    & \multicolumn{2}{c}{Bedrooms}  &      \multicolumn{2}{c}{Living Rooms}  \\
\cline{2-5}
         & $\#$Rel. & $\#$Obj. & $\#$Rel. & $\#$Obj.  \\
\hline

HOLODECK & 0.77 & 0.93 &  0.69 & 0.97 \\
\hline

Our & \textbf{0.88} & \textbf{0.99} & \textbf{0.86} &  \textbf{1.00} \\
\hline
\end{tabular}
\caption{Semantic alignment between LLM-generated descriptions and the resulting scenes of HOLODECK and ours.}
\label{tab:comparison_holo}
\end{table}

We further provide an additional quantitative comparison with HOLODECK, the closest work to ours, as both use the LLM to generate textual scene descriptions and then solve for the scene layouts. The difference is that HOLODECK requires dense and detailed spatial relations while ours uses hierarchical structures with sparse relations as well as a neural network to infer the fine-grained relative placements. Table~\ref{tab:comparison_holo} reports the semantic alignment between the LLM-generated descriptions and the generated scenes. Specifically, $\#$Rel. counts the percentage of relative placements matching with LLM-generated spatial relations and $\#$Obj. counts the existence of LLM-specified objects. Obviously, our results align better with LLM arrangements, implying the advantages of using hierarchical scene representation with our approach.

\begin{table}[]
\centering
\setlength{\tabcolsep}{7pt}
\begin{tabular}{ccccc}
\hline
Models     &  Scene Effect. & Physical. & Layout. \\
\hline
ATISS      & 3.55           & 2.86              & 3.10       \\
\hline
DiffuScene & 3.38           & 2.55              & 2.57      \\
\hline
LayoutGPT  & 3.04          & 2.45           & 2.23       \\
\hline
Holodeck   & 3.12          & 3.75             & 3.23       \\
\hline
Ours       & \textbf{4.73}          & \textbf{4.69}            & \textbf{4.55}  \\
\hline
\end{tabular}
\caption{Averaged score of perceptual study on the generated scenes. The participants give scores between 1-5 w.r.t. scene effectiveness, physical feasibility and layout rationality.}
\label{tab:perceptual_study}
\end{table}

\noindent\textbf{Perceptual Study.} The perceptual study evaluates the quality of scenes generated by different methods (those used in the comparison experiments). We present 25 generated scenes with the input user requirements to 30 participants, who rate them on a 5-point Likert scale from three aspects: scene effectiveness, physical feasibility, and layout rationality. Since both the rendered scenes and the topview visualizations are presented, the participants are sensitive to unreasonable placements which might be covered by the occlusion in the renderings. Table~\ref{tab:perceptual_study} shows that our results get the highest scores w.r.t. all three aspects, i.e. high scene effectiveness as we prompt the LLM with functional area considerations, high feasibility and layout rationality as our approach can effectively generate feasible scenes aligned with LLM arrangement. Otherwise, although HOLODECK prevents object overlap or out-of-boundary, since its results may break the LLM arrangements, it may affect possible human activity, such as shown in the top row case in Figure~\ref{fig:comparison_topview}.

\begin{figure}
\centering
\includegraphics[width=1\linewidth]{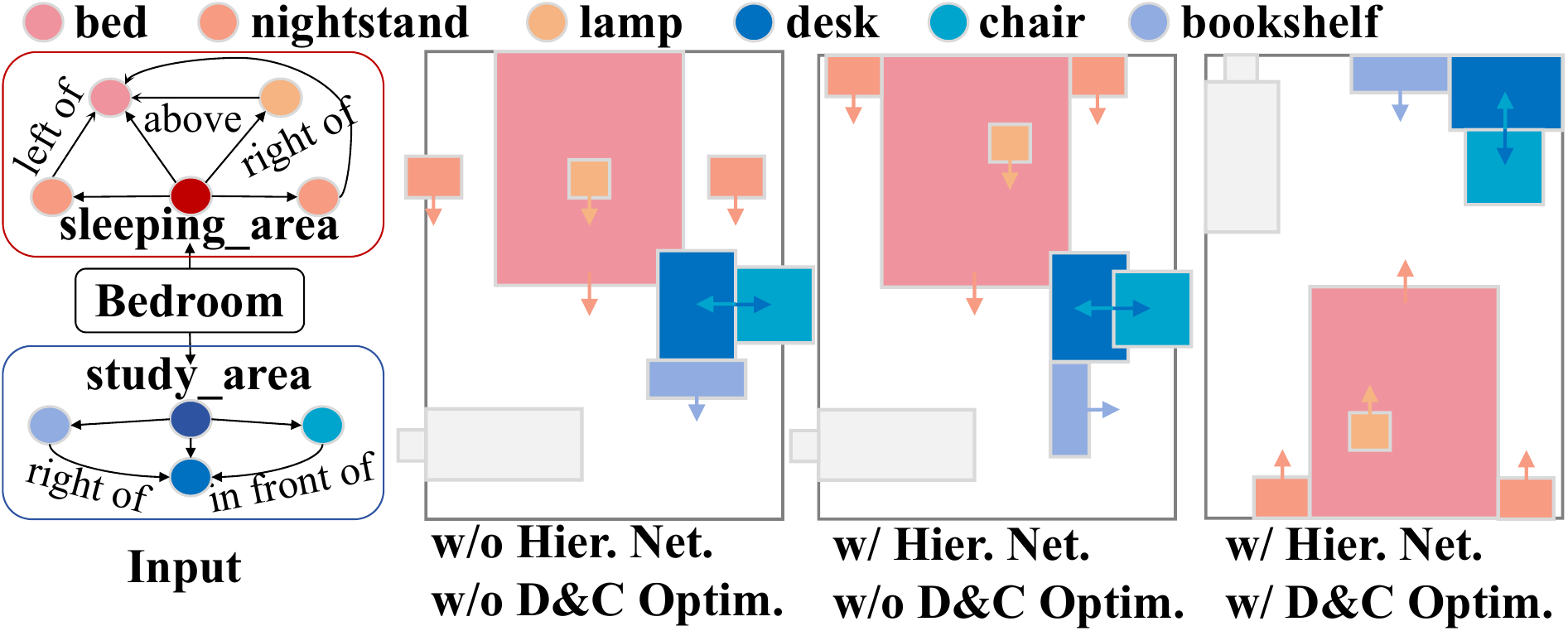}
\caption{Topview visualizations of the ablation study.}
\label{fig:ablation_topview}
\end{figure}
\begin{table*}
\setlength{\tabcolsep}{10pt}
\centering

\begin{tabular}{cc|ccc|ccc}
\hline
\multicolumn{2}{c}{Baselines} & \multicolumn{3}{c}{Bedrooms} & \multicolumn{3}{c}{Living Rooms} \\

\hline
Hierarchy-Aware Net. & D$\&$C Optim. & {OOB↓} & {Overlap↓} & {KL Div.↓}  & {OOB↓} & {Overlap↓} & {KL Div.↓} \\
\hline

$\times$ & $\times$ &  0.80 & 0.09 & 0.23 & 0.98 & 0.10 & 0.24  \\
\hline

\checkmark & $\times$ &  0.73&  0.18&  0.09&  0.93&  0.21& 0.16  \\\hline

$\times$ & \checkmark & 0.00 & 0.00 & 0.23 & 0.00 & 0.00 & 0.23  \\\hline

\checkmark & \checkmark &  \textbf{0.00}& \textbf{0.00}& \textbf{0.09}&  \textbf{0.00}&  \textbf{0.00}& \textbf{0.13} \\ 
\hline

\end{tabular}
\caption{Quantitative evaluation of ablation study to validate our key designs: the hierarchy-aware network (Hierarchy-Aware Net.) to infer fine-grained relative placements and divide-and-conquer optimization (D$\&$C optim.) to solve the final layouts.}
\label{tab:ablation_study}
\end{table*}

\subsection{Ablation Study}

The ablation study validates the key designs of our approach, i.e. the hierarchy-aware network and the divide-and-conquer optimization, by removing the corresponding stage from our pipeline. When removing the hierarchy-aware network, we pre-define the relative placement coordinates for the textual spatial relations to replace the predictions of the network. When removing the divide-and-conquer optimization, we directly require the LLM to generate coordinates of anchor objects and transform the relative placements of the other objects into global coordinates without optimization refinement. The results are shown in Table~\ref{tab:ablation_study} and Figure~\ref{fig:ablation_topview}. It indicates that our hierarchy-aware network is of vital importance in capturing the reasonable relative placements between objects, and the divide-and-conquer optimization ensures the physical feasibility of the generated scene layouts.

\begin{figure}
\centering
\includegraphics[width=\linewidth]{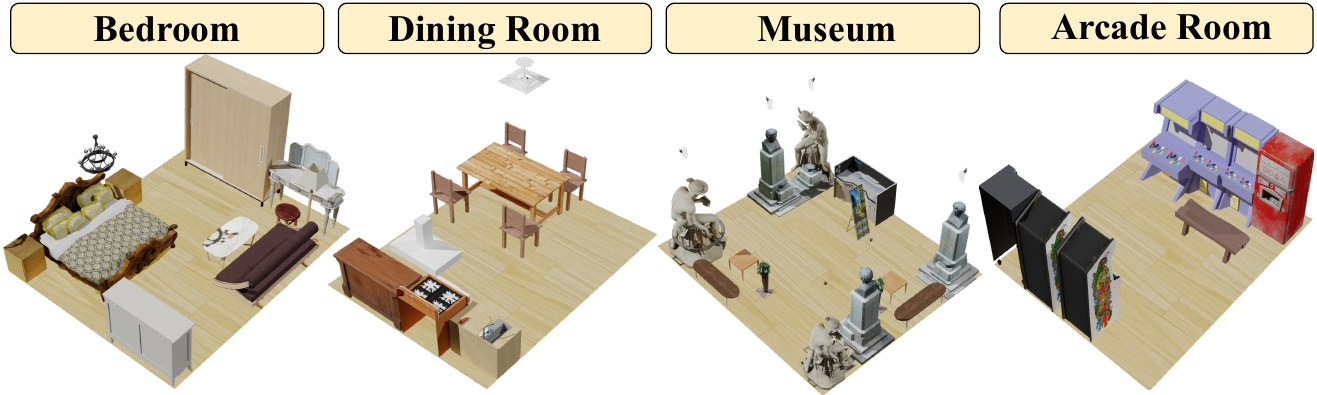}
\caption{Diverse results in the open-vocabulary task.}
\label{scene_diversity}
\end{figure}

\section{Applications}

\noindent\textbf{Open-vocabulary scene synthesis.} To demonstrate the generalization of our approach, we show the open-vocabulary scene synthesis results in Figure~\ref{scene_diversity}. By removing object category constraints in the LLM prompt, we generate scenes of a dining room, museum, and arcade room, not trained on by our model, to validate its practical usage. The results show that the LLM generates reasonable hierarchical structures for arbitrary requirements, while our generalizable hierarchy-aware network and divide-and-conquer optimization produce realistic scenes with various descriptions.

\begin{figure}[t]
\centering
\includegraphics[width=\linewidth]{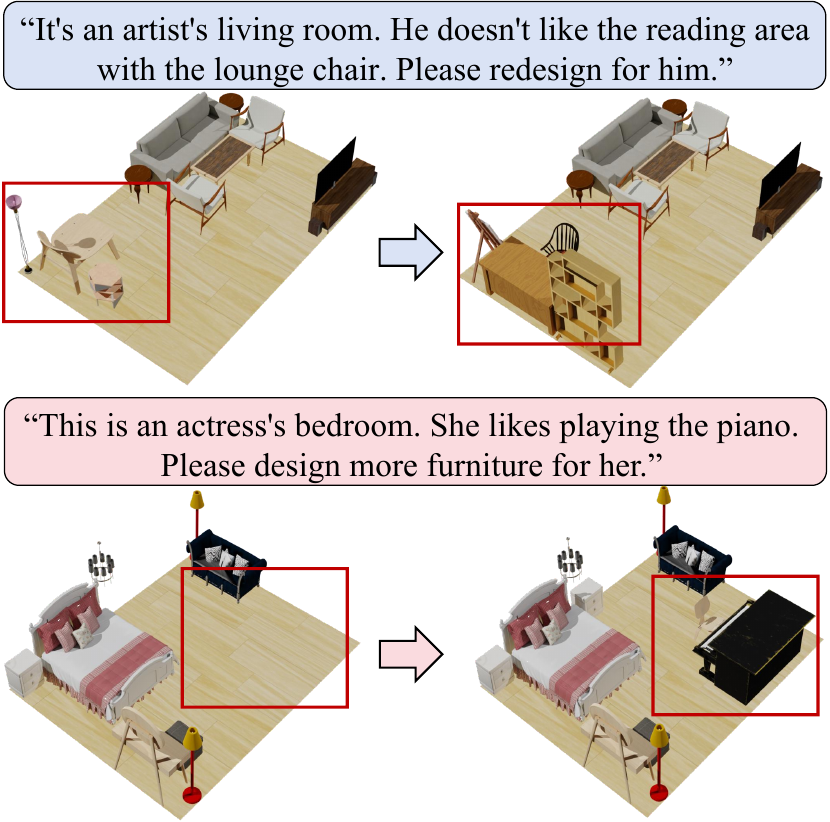}
\caption{Language-guided interactive scene editing.}
\label{scene_editing}
\end{figure}

\noindent\textbf{Interactive scene editing.} Our approach also supports user-friendly language-guided interactive scene editing. Specifically, we describe the current state of the scene and an editing instruction as the input to LLM, and add an additional constraint to the divide-and-conquer optimization to maintain the placements of unchanged objects as much as possible. As shown in Figure~\ref{scene_editing}, the LLM can modify the scene by adding and removing objects. Moreover, with the hierarchical scene structure and our approach, the edited scenes exhibit minimal changes from the original scenes while satisfying the LLM arrangments and the expectations of users.
\section{Conclusions}

We present an LLM-assisted hierarchical indoor scene synthesis approach to produce customized and diverse scenes. Our approach fully takes advantage of the three-level hierarchical structure, where the LLM generates the descriptions of hierarchical scenes, a hierarchy-aware network infers the fine-grained relative placements, and a divide-and-conquer optimization solves the feasible layout.

Our approach still holds some limitations. First, for the simplicity of optimization, we assume rectangular floors for the generated scene. It is possible to utilize the spatial-aware optimization with simulated annealing algorithms~\cite{Yu2011MakeIH} for irregular floors. Second, LLM sometimes generates infeasible configurations with too many or large objects and we randomly remove some areas or objects to address this, affecting scene quality. Third, since our hierarchical scene representation takes each object as its oriented bounding boxes without geometric details, our generated scenes are not sensitive to object shapes, such as L-shaped sofa. 

\section*{Acknowledgments}

This work is supported in part by the Shandong Province Excellent Young Scientists Fund Program (Overseas) (Grant No. 2022HWYQ-048 and 2023HWYQ-034), the TaiShan Scholars Program (Grant No. tsqn202211289), the National Natural Science Foundation of China (Grant No. 62325211 and 62132021).

\bibliography{references}

\end{document}